# APLICAÇÃO DE ROS COMO FERRAMENTA DE ENSINO A ROBÓTICA

*Daniel Maia Evangelista – danielmenator@gmail.com*
*Graduando em Engenharia de Controle e Automação na Universidade de Fortaleza*
*Endereço: Av. Washington Soares, 1321 – Edison Queiroz, Fortaleza – CE, Brasil*

*Pedro Benevides Cavalcante – pedrobenevides3243@gmail.com*
*Graduando em Engenharia de Controle e Automação na Universidade de Fortaleza*
*Endereço: Av. Washington Soares, 1321 – Edison Queiroz, Fortaleza – CE, Brasil*

*Afonso Henriques Fontes Neto Segundo – afonsof@unifor.br*
*Professor na Universidade de Fortaleza (UNIFOR)*
*Endereço: Av. Washington Soares, 1321 – Edison Queiroz, Fortaleza – CE, Brasil*

**Resumo:** *O estudo dos manipuladores robóticos é o objetivo principal da disciplina de Robótica Industrial presente no curso de formação do engenheiro de controle e automação. Existe a dificuldade de preparar práticas e projetos acadêmicos na área da robótica devido ao elevado custo dos equipamentos educacionais específicos. Assim, como as aulas práticas e o desenvolvimento de projetos são de grande importância para a formação do engenheiro, propõe-se usar um software de simulação a fim de se proporcionar uma experiência pratica para os alunos da disciplina. Nesse contexto, o presente artigo visa expor a utilização do Robot Operation System (ROS) como ferramenta para desenvolver um braço robótico virtual e implementar a funcionalidade das cinemáticas direta e inversa, estudadas na disciplina de Robótica Industrial. Tal desenvolvimento poderá ser utilizado como ferramenta educacional, para aumentar o interesse e o aprendizado dos alunos da disciplina de robótica e ampliar áreas de pesquisa para a disciplina.*

**Palavras-chave:** *Robótica. ROS. Aprendizagem baseada em Projetos. Engenharia de Controle e Automação.*

## 1    INTRODUÇÃO

Existem inúmeros desafios na formação profissional dos engenheiros. De acordo com Rapkiewicz et. al. (2006), uma das grandes dificuldades dos acadêmicos ingressantes nas graduações de engenharia é aplicar o conhecimento teórico e visualizar situações práticas na sua realidade profissional. Tais dificuldades implicam na desmotivação do indivíduo.

A Aprendizagem Baseada em Projeto (ABP ou *PBL* em inglês – *Project Based Learning*) proporciona que o aluno seja exposto a desafios da realidade profissional e aprimore suas habilidades de resolução de problemas (BARREL, 2010). Projetos, assim como aulas práticas, influenciam o entendimento da matéria e servem como meio de os alunos aplicarem assuntos complexos e avançados da área (SUGAYA; OHBA; KANMACHI, 2017).

Este artigo tem como objetivo propor um projeto para a disciplina de Robótica, presente no curso de Engenharia de Controle e Automação, que busca proporcionar para os alunos uma oportunidade de aplicar e visualizar os conteúdos estudados na disciplina, e também pode gerar a oportunidade de aplicar outras matérias significativas para a formação do engenheiro.

Para alcançar a proposta do projeto, o *Robot Operation System* (*ROS*) ou em português, Sistema Operacional de Robô, será utilizado. O *ROS* consiste em um open-software que provê toda estrutura computacional para a implementação de programas para controlar qualquer tipo de robô e também ferramentas de simulação do mundo real, ideais para a realização de testes anteriores à construção ou montagem do modelo físico (AITKEN, VERES, JUDGE, 2014). A utilização do *ROS* possibilita que qualquer universidade possa aplicar o projeto descrito, sem custos adicionais na disciplina.

## 2  ROS

O *ROS* é uma estrutura (*Framework*) flexível de código aberto para desenvolver *softwares* para robôs. É uma coleção de ferramentas, bibliotecas e convenções que visam facilitar tarefas de criação e desenvolvimento de robôs complexos e robustos para serem utilizados nas mais diversas plataformas robóticas. Ele fornece os serviços normalmente utilizados em um sistema operacional, como abstração de hardware, controle de dispositivo de baixo nível, implementação de funcionalidade comumente usada, passagem de mensagens entre processos e gerenciamento de pacotes (QUIGLEY et al., 2009). Ele também fornece ferramentas e bibliotecas para obter, criar, gravar e executar código em vários computadores.

O principal recurso do *ROS* é a forma como o *software* é executado e a maneira como ele se comunica, permitindo que se projete software complexo sem saber como determinado *hardware* funciona. O *ROS* fornece uma maneira de conectar uma rede de processos (nós, em inglês *nodes*) a um *hub* central. Os nós podem ser executados em vários dispositivos e se conectam a esse *hub* na topologia de ponto a ponto (QUIGLEY et al., 2009).

As principais formas de criar a rede são fornecer serviços solicitáveis ou definir conexões de *publisher* / *subscriber* (publicador / assinante) com outros nós. Ambos os métodos se comunicam por meio de tipos de mensagem especificados. Alguns tipos são fornecidos pelos pacotes principais, mas os tipos de mensagens podem ser definidos por pacotes individuais (QUIGLEY et al., 2009).

## 3  GAZEBO

O *Gazebo* é um simulador projetado para reproduzir com precisão os ambientes dinâmicos que o robô possa encontrar. Essa ferramenta modela ambientes externos e fornece um *feedback* realista dos sensores para o robô e o usuário (KOENIG; HOWARD, 2004).

Objetos com massa, velocidade, fricção e outros atributos podem ser simulados no *Gazebo*, realizando trabalhos de empuxo, carregamento e até à força da gravidade de forma realista. Essas ações podem ser usadas como parte integrante de uma experiência (KOENIG; HOWARD, 2004).

Forças angulares e lineares podem ser aplicadas em um braço robótico dentro do simulador Gazebo, gerando locomoção e interações com o ambiente. Ele permite que praticamente todos as configurações sejam editadas, podendo assim oferecer ao usuário uma ampla possibilidade de simulações (KOENIG; HOWARD, 2004).

O *Gazebo*, assim como o ROS, é um *software* totalmente *open-source*, o que é uma grande vantagem em frente aos outros *softwares* do mercado. Além disso, ele possui uma ampla comunidade de contribuintes que estão em rápida evolução para atender as mudanças necessárias (KOENIG; HOWARD, 2004).

Esse incrível software é uma excelente ferramenta para desenvolvimento e testes de robôs, podendo ser usado também em pesquisas no campo da robótica, por isso à sua crescente utilização (KOENIG; HOWARD, 2004).

# 4 PROCEDIMENTO DO PROJETO

O projeto proposto consiste em cinco etapas. A primeira consiste na modelagem física do manipulador robótico, esta é a etapa que mais foge da matéria abordada em Robótica Industrial, mas aborda assunto de outras disciplinas presentes na formação do engenheiro de controle (cadeiras que abordam CAD, ex: Projetos Assistidos por Computadores e Desenho para Engenharia, na grade da Universidade de Fortaleza). A segunda é o desenvolvimento dos cálculos das cinemáticas (inversa e direta) que envolvem o manipulador modelado, a terceira etapa é a modelagem e a simulação virtual do modelo físico no ROS. A quarta etapa é a elaboração de algoritmos para a realização das cinemáticas anteriormente calculadas e, por último a demonstração e explicação de o que e como realizou cada etapa anterior.

## 4.1 Desenvolvimento do Braço

A criação do braço para a simulação não é o foco deste trabalho, mas como é essencial que se haja um manipulador robótico modelado para o desenvolvimento do, será apresentado resumidamente o necessário.

### 4.1.1. Modelagem CAD

Na primeira parte do projeto deverá ser desenhado em qualquer software CAD, o manipulador que será controlado no projeto. O desenho do manipulador não pode ser fixo e sim articulado para a matéria poder ser aplicada. Após o desenho do manipulador deve se estabelecer um material para o braço e assim, com ajuda do software CAD obter a massa e os referenciais inerciais de cada parta constituinte do manipulador. Para finalizar, os desenhos devem ser exportado no formato .STL, compatível com a ferramenta *Gazebo*.

O objetivo desta sessão é obter o desenho no formato compatível e os parâmetros definidores do manipulador. Estes são: os números de articulações, os tamanhos das ligações, os pesos de cada parte e as referências inerciais.

Figura 1 – Exemplo de Parte CAD em STL

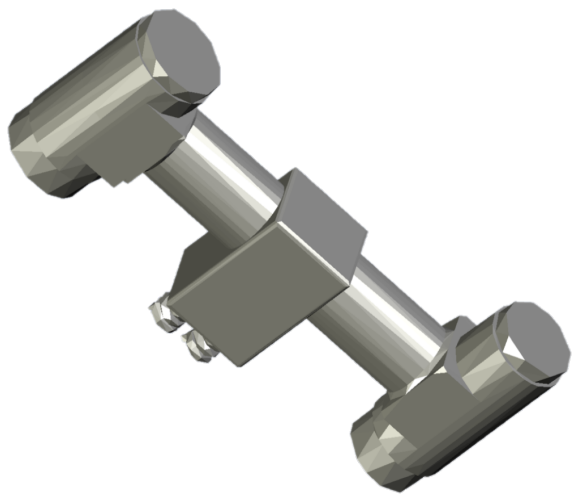

Fonte: Elaborado pelo Autor

Nessa primeira etapa, também é possível se obter o modelo CAD em sites na internet. Porem nesse caso torna-se mais difícil se obter os parâmetros definidores que são essenciais para o progresso do projeto. Um exemplo de modelos que pode se obter é o *katana_arm_gazebo*.

### 4.1.2. Modelagem URDF

A modelagem realizada em URDF, consiste na representação das partes e dos eixos de rotação de um modelo CAD para a ferramenta Gazebo. Ela é composta em linhas de código XML contendo os parâmetros que definem cada parte: massa, referencias inerciais e forma geométrica. Para a forma geométrica recomendasse a utilização o arquivo .STL, gerado na modelagem CAD, assim é obtida uma representação mais realista, mas também pode ser usadas representações de figuras geométricas básicas, como paralelepípedos e cilindros.

Para que o modelo gerado possa ser controlado, conforme o desejado, além da modelagem do manipulador deve-se adicionar ao arquivo URDF o *plugin* chamado de *gazebo_ros_control* e declarar transmissões para cada eixo de rotação. As transmissões, são as responsáveis por movimentar os eixos, elas exigem a declaração de uma interface física que modifica o seu funcionamento, para a aplicação a *EffortJointInterface* é a mais interessante, esta recebe um valor de ângulo e posiciona o eixo de rotação neste.

Figura 2 – Exemplo de Arquivo URDF

```
<!-- BGN - Robot description -->
<m_link_box name="${link_00_name}"
        origin_rpy="0 0 0" origin_xyz="0 0 0"
        mass="1024"
        ixx="170.667" ixy="0" ixz="0"
        iyy="170.667" iyz="0"
        izz="170.667"
        size="1 1 1" />

<m_joint name="${link_00_name}__${link_01_name}" type="revolute"
        axis_xyz="0 0 1"
        origin_rpy="0 0 0" origin_xyz="0 0 0.5"
        parent="base_link" child="link_01"
        limit_e="1000" limit_l="-3.14" limit_u="3.14" limit_v="0.5" />

<m_link_mesh name="${link_01_name}"
        origin_rpy="0 0 0" origin_xyz="0 0 -0.1"
        mass="157.633"
        ixx="13.235" ixy="0" ixz="0"
        iyy="13.235" iyz="0"
        izz="9.655"
        meshfile="package://mrm_description/meshes/Link1-v2.stl"
        meshscale="0.001 0.001 0.001" />
```

Fonte: Elaborado pelo Autor

## 4.2 Cinemática

Em robótica, cinemática se refere ao estudo da movimentação dos corpos constituintes do robô, desconsiderando forças e massas atuantes. A partir do estudo da cinemática de um manipulador robótico, o movimento de todas as articulações que o compõem e o movimento de sua ferramenta para a execução de tarefas podem ser previstos. A equação de cinemática direta determina a posição terminal do robô, se todas as variáveis articulares forem conhecidas. O estudo de cinemática inversa permite calcular qual deve ser o valor de cada variável articular, a partir de uma determinada coordenada final (NIKU, 2011).

A cinemática direta ou equações de configuração cinemática, para robôs rígidos, determina a relação entre os eixos constituintes do braço e também a posição e a orientação da ferramenta do manipulador (SPONG; HUTCHINSON; VIDYASAGAR, 2008). Isso indica que para a análise cinemática é necessário que se tenha anteriormente o braço modelado. Utilizando a representação de Denavit Hartenberg obtém-se a análise representante da cinemática direta (KUCUK; BINGUL, 2007).

A cinemática inversa é essencial para o controle de manipuladores robóticos e é um problema estudado há muitas décadas. Há dois meios para se encontrar a solução para o problema, o método geométrico e o método algébrico. O método geométrico é geralmente usado para manipuladores simples como braços de até 3 graus de liberdade (em inglês, *DOF*, *degrees of freedom*), como o braço-exemplo. Para manipuladores mais complexos usa-se o método algébrico, que depende da cinemática direta (KUCUK; BINGUL, 2007).

Ao contrário da cinemática direta, que é definidora e única, a cinemática inversa pode gerar múltiplas soluções e várias singularidades a serem observadas. Assim, toda vez que forem calculados os ângulos articulados a partir da cinemática inversa, deve-se verificar se o manipulador atingirá o valor desejado a partir da cinemática direta (KUCUK; BINGUL, 2007).

### 4.3 Desenvolvimento da Simulação

Nesta sessão será explicado de maneira breve como deve proceder o desenvolvimento da simulação. Como esta sessão depende do ROS que não é um sistema simples e que passa por constantes alterações, uma explicação detalhada torna-se longa e vulnerável a erros. Qualquer problema encontrado no desenvolvimento em relação ao ROS, pode-se procurar solução em sua extensa documentação ou fazer uma pergunta na página de sua ativa comunidade.

Dentro do ambiente do ROS, um pacote, onde todo o projeto será desenvolvido, deve ser criado e dentro dele um arquivo executável (*roslaunch*), para começar a implementação da simulação. Neste arquivo executável será declarada a iniciação do *Gazebo* e do nó *urdf_spawner*, responsável por a partir de um URDF gerar o manipulador dentro do ambiente do Gazebo. Quando o braço é gerado dentro *Gazebo* com sucesso, o *urdf_spawner* tenta também gerar nodes responsáveis por controlar a posição de cada articulação declarada no URDF, mas para esta operação ser bem-sucedida é necessária a inclusão de controladores para cada articulação e das transmissões no URDF, citado na Seção 4.1.2. A Figura 3 apresenta um exemplo de controlador declarado.

Figura 2 – Exemplo de Controlador para Eixos de Rotação

```
arm_model:
  # Publish all joint states -----------------------------------
  joint_state_controller:
    type: joint_state_controller/JointStateController
    publish_rate: 50

  # Position Controllers --------------------------------------
  joint1_position_controller:
    type: effort_controllers/JointPositionController
    joint: base_to_00
    pid: {p: 100.00, i: 0.01, d: 10.00}

  joint2_position_controller:
    type: effort_controllers/JointPositionController
    joint: 00_to_01
    pid: {p: 100.00, i: 0.01, d: 10.00}

  joint3_position_controller:
    type: effort_controllers/JointPositionController
    joint: 01_to_02
    pid: {p: 100.00, i: 0.01, d: 10.00}
```

Fonte: Elaborado pelo Autor

Para a simulação ter sucesso na execução tem que se prestar atenção em diversos detalhes, como por exemplo nome de variáveis que não podem ser mudadas, disposição de pastas, entre outros. Se não for o objetivo que o aluno tenha que ir atrás de aprender a manusear a ferramenta ROS, pode ser provido o pacote já previamente preparado, com a simulação já bem-sucedida e o aluno terá apenas o desafio de modelar seu manipulador e realizar os algoritmos estudados na disciplina, já que o aprendizado de ROS não consta na ementa. Uma vez, a simulação executada com sucesso dificilmente acontecerá algo que venha a causar problemas.

Figura 3 – Simulação de Manipulador no Gazebo

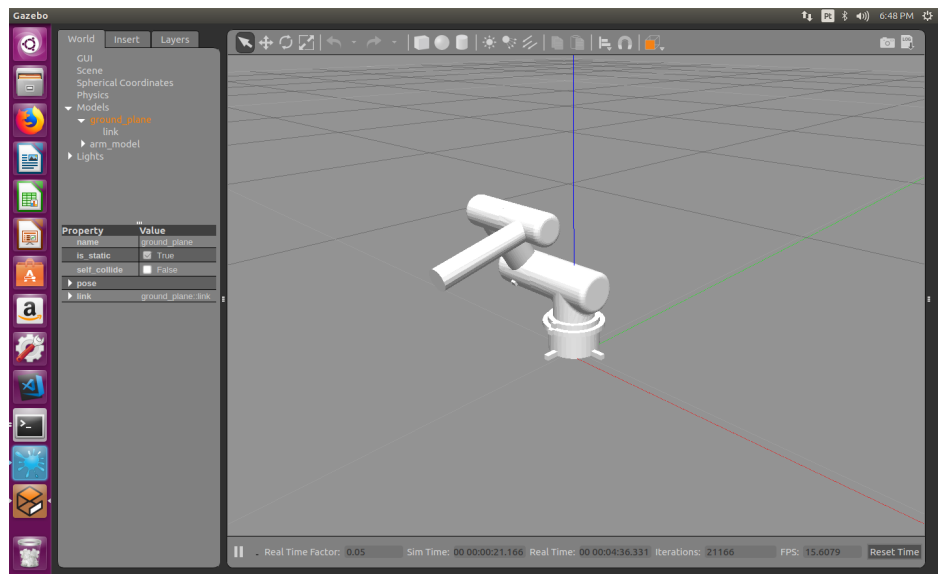

Fonte: Elaborado pelo Autor

### 4.4 Implementação de Cinemáticas

Com os cálculos das cinemáticas e com a simulação funcionando, só resta implementar algoritmos que controlem a simulação, para finalizar o projeto. Para a execução das cinemáticas é recomendado que se crie um *Node Subscriber*, este pode ser escrito em *Python* ou *C++* e poderá ser chamado por outros Nós, se for desejado. O *Node* criado pode receber parâmetros iniciais como funções de programas, então uma maneira de se implementar os *Nodes* que realizem as cinemáticas direta e indireta é fazer um que receba os valores dos ângulos desejados, para a direta e um que receba os valores de posição desejada, para a inversa.

Para a cinemática direta a implementação em algoritmo fica simples, tanto em *Python* quanto em *C++*. Após receber o valor dos ângulos desejados, basta que o Subscriber desenvolvido publique nos *Nodes* responsáveis pelo controle das articulações os seus respectivos valores. Para garantir que a operação foi bem-sucedida e que está condizendo com os cálculos previamente realizados, pode-se verificar se as posições dos eixos pelos mesmos nós responsável responsáveis pelo controle das articulações.

Com a resolução da cinemática inversa previamente resolvida a implementação em algoritmo também fica simples. Para encontrar os ângulos necessários para chegar na coordenada desejada, basta realizar as contas que representam a cinemática inversa do manipulador. Porem de acordo com Kucuk e Bingul (2006), devido as singularidades da cinemática inversa é necessário que se verifique se os cálculos condizem com o resultado da cinemática direta. Se for confirmado que o manipulador atingirá o valor desejado, de mesma forma que é realizado no algoritmo da cinemática direta, basta que o Subscriber desenvolvido publique os valores de angulação encontrado.

# 5 CONSIDERAÇÕES FINAIS

A utilização de projetos na assimilação de competências nos cursos de engenharia é algo fundamental para a formação de profissionais, que contribui no aprendizado acadêmico e científico. Nesse contexto, o projeto a qual este artigo se refere comprova-se como ponto positivo. Observa-se que o projeto exige domínio dos assuntos abordados na disciplina de Robótica, então poderia ser aplicado como ferramenta educacional.

Utilizando o ROS o projeto ganha além da importância educacional ganha também a importância tecnológica, já que é uma plataforma com uma crescente relevância prevista. Mas a principal relevância que a aplicação desse projeto traz é a redução de custo para se aplicar práticas de robótica. Quigley, Asbeck e Ng (2011) expõe um projeto de manipulador robótico de baixo custo que custa $4135, valor ainda considerado alto por muitos.

Devido as possibilidades de se aplicar mais conteúdo como por exemplo: cinemática jacobiana e planejamento de trajetória planeja-se aplicar esse projeto em uma turma da disciplina, para se obter resultados sobre o aprendizado que este pode transmitir.

# USING ROS AS A ROBOTICS TEACHING TOOL

***Abstract:*** *The study of robotic manipulators is the main goal of Industrial Robotics Class, part of Control Engineers training course. There is a difficulty in preparing academic practices and projects in the area of robotics due to the high cost of specific educational equipment. The practical classes and the development of projects are very important for engineers training, it is proposed to use simulation software in order to provide practical experience for the students of the discipline. In this context, the present article aims to expose the use of the Robot Operation System (ROS) as a tool to develop a robotic arm and implement the functionality of forward and inverse kinematics. Such development could be used as an educational tool to increase the interest and learning of students in the robotics discipline and to expand research areas for the discipline.*

***Key-words:*** *Control Engineering, robotic, ROS.*